\documentclass[letterpaper, 10 pt, conference]{ieeeconf}

\IEEEoverridecommandlockouts
\usepackage{subcaption}
\usepackage{amsmath}
\usepackage{amssymb}
\usepackage{booktabs}
\usepackage{multirow}
\usepackage{tabularx}
\usepackage{xcolor}
\usepackage{url}
\usepackage{cite}
\usepackage{tabularx}
\usepackage{array}
\newcolumntype{Y}{>{\centering\arraybackslash}X}
\usepackage{algorithm}
\usepackage{algorithmic}
\usepackage{graphicx}

\newcommand{\method}{\textsc{PFD}}

\title{\LARGE \bf
Perturbation-Based Epistemic Uncertainty for Failure Detection in Vision-Language-Action Models
}

\author{Yousung Lee and Dongsoo Har%
\thanks{The authors are with Korea Advanced Institute of
Science and Technology (KAIST), Daejeon, Republic of Korea.}%
}

\begin{document}

\maketitle
\thispagestyle{empty}
\pagestyle{empty}

\begin{abstract}

Vision-Language-Action (VLA) models have shown strong performance in robotic manipulation, but reliable uncertainty quantification remains challenging, particularly under distribution shift.
Unlike autoregressive policies, many modern VLA models generate continuous actions through regression or flow-based generation, where explicit predictive probabilities are unavailable.
Moreover, stochastic action sampling primarily captures action-generation variability under a fixed model, while failure detection under distribution shift can benefit from capturing uncertainty in the model itself.
Motivated by Bayesian perspectives on local model variations,
we propose perturbation-based failure detection
(\method), a training-free framework for estimating epistemic
uncertainty in VLA models through low-rank weight perturbations.
Specifically, we inject random low-rank perturbations into
selected transformer weight matrices and estimate epistemic
uncertainty from disagreement across perturbed action predictions.
Experiments on LIBERO-PRO show that \method{} achieves the
highest average AUROC and balanced accuracy among the evaluated
methods while consistently outperforming stochastic action sampling
across distribution shifts.
Real-world robot experiments further demonstrate that \method{}
provides a competitive failure-detection signal under an unseen
object shift.

\end{abstract}

\section{Introduction}
Vision-Language-Action (VLA) models have recently demonstrated
strong generalization capabilities across robotic manipulation
tasks by mapping multimodal observations and language
instructions to continuous control actions
~\cite{kim2024openvla, kim2025oft, black2025pi05}.
However, despite their strong performance, VLA systems can fail when they encounter situations that deviate from the training distribution. 
This is particularly concerning in real-world robotics, where even small action errors can lead to task failures or unsafe robot behavior.
Recent studies have shown that VLA models can exhibit
overconfident failures under distribution shift, sometimes
ignoring language instructions and relying on memorized or
shortcut-like manipulation patterns~\cite{zhou2025liberopro, xing2025shortcut}.
Such failures are problematic because the model may continue executing incorrect actions despite violating the language instruction.

Reliable uncertainty estimation is therefore critical for safe
deployment and failure detection in VLA systems.
However, uncertainty estimation in VLA models is fundamentally
challenging.
Unlike autoregressive token-based VLA models
~\cite{brohan2023rt2,kim2024openvla}, many recent VLA models
generate actions through regression or flow-based generation
~\cite{kim2025oft,black2025pi05}, where explicit predictive
probabilities are unavailable.
As a result, uncertainty estimation methods based on token-level entropy or perplexity in autoregressive VLA policies~\cite{karli2025ask} are not directly applicable to these models.

From a Bayesian perspective, predictive uncertainty is commonly
decomposed into aleatoric uncertainty and epistemic uncertainty.
Aleatoric uncertainty captures inherent stochasticity in observations or action generation and is generally irreducible, whereas epistemic uncertainty reflects uncertainty over model parameters due to limited or insufficient training data~\cite{kendall2017uncertainties,gal2016dropout,lakshminarayanan2017deep}.
In generative policies, stochastic sampling is a natural way to estimate uncertainty by measuring variability across multiple action samples~\cite{roemer2025fiper}.
For flow-based models such as $\pi_{0.5}$~\cite{black2025pi05}, varying the initial action-generation noise enables the policy to represent multiple plausible action modes for the same observation.
The resulting action variability therefore primarily reflects
aleatoric uncertainty under a fixed pre-trained model.
In safety-critical robotic systems, epistemic uncertainty is
particularly important under distribution shift, where policies
may encounter situations that are poorly represented by their
training data.
A reliable policy should therefore provide an uncertainty signal indicating when its behavior is poorly constrained by the training data and, consequently, less reliable.

\begin{figure}[t]
    \centering
    \includegraphics[width=\linewidth]{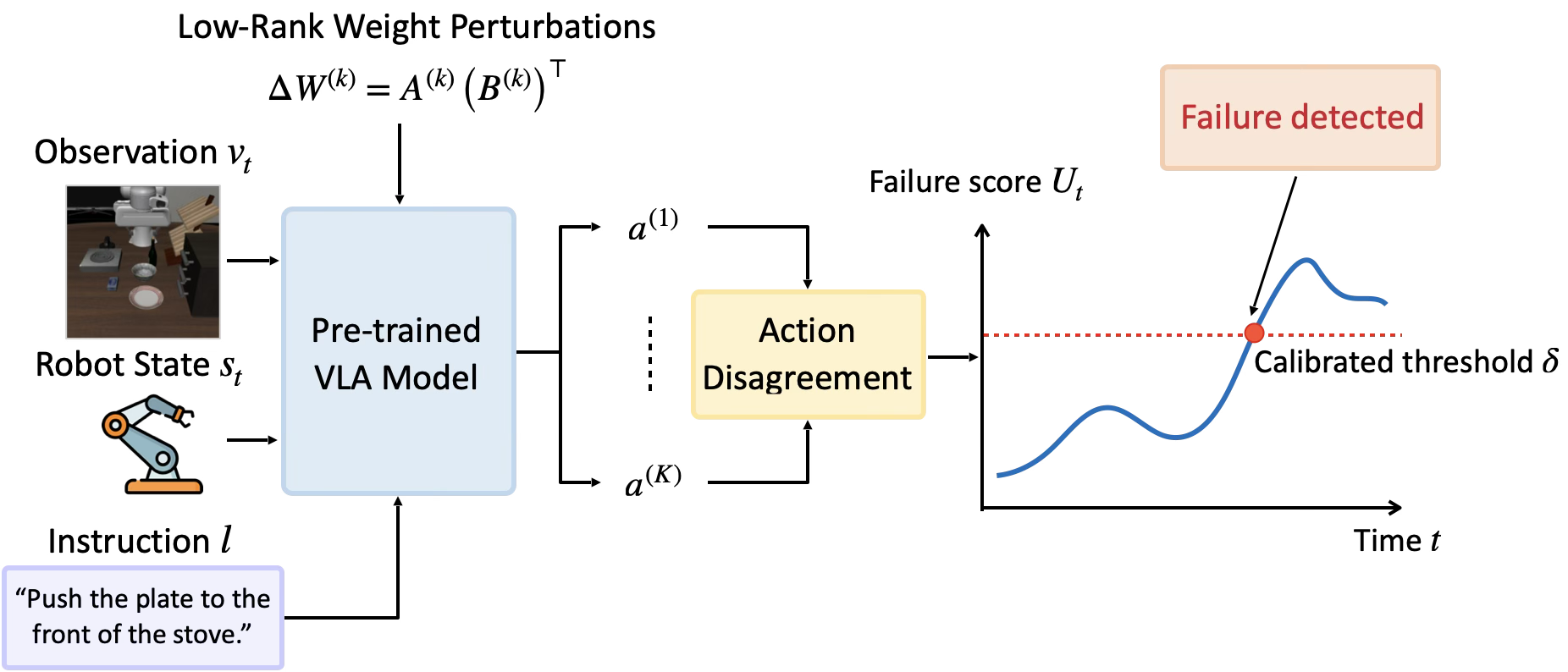}
    \caption{
    Overview of the proposed \method{} framework.
    Given the current observation, robot state, and language
    instruction, low-rank weight perturbations generate local parameter
    variations around the pretrained VLA model.
    Disagreement among their action predictions provides an
    epistemic uncertainty-based failure score $U_t$.
    A failure is detected when $U_t$ exceeds the calibrated
    threshold $\delta$.
    }
    \label{fig:method}
    \vspace{-2mm}
\end{figure}

\begin{figure*}[t]
    \centering
    \begin{minipage}[t]{0.495\textwidth}
        \centering
        \includegraphics[width=\linewidth]{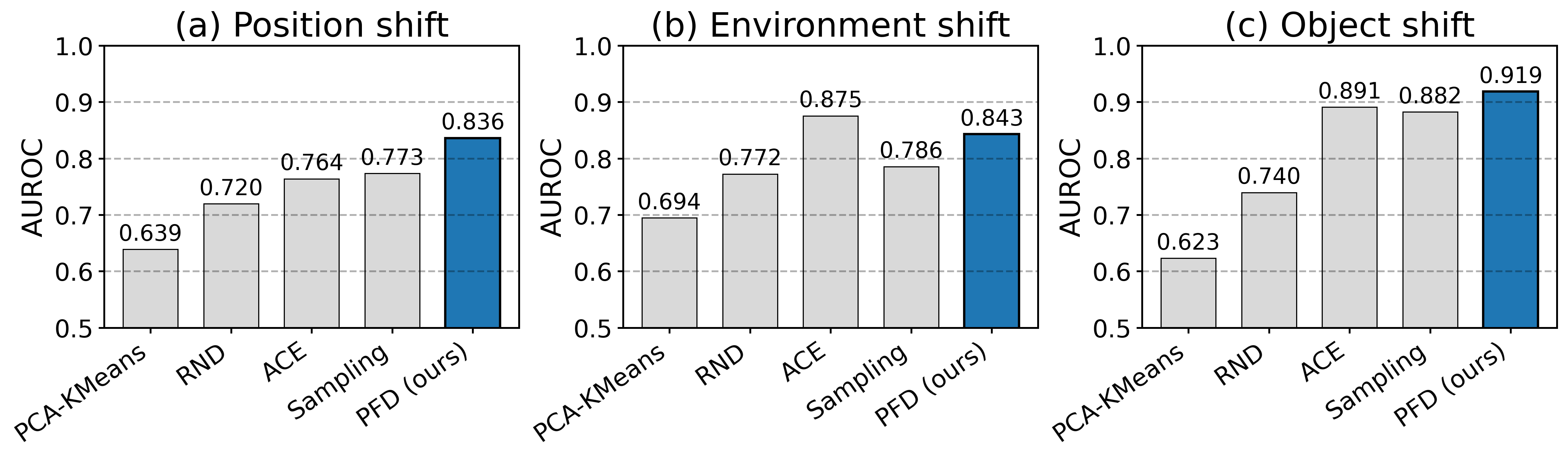}\\[-1mm]
        \small (a) AUROC
    \end{minipage}
    \hfill
    \begin{minipage}[t]{0.495\textwidth}
        \centering
        \includegraphics[width=\linewidth]{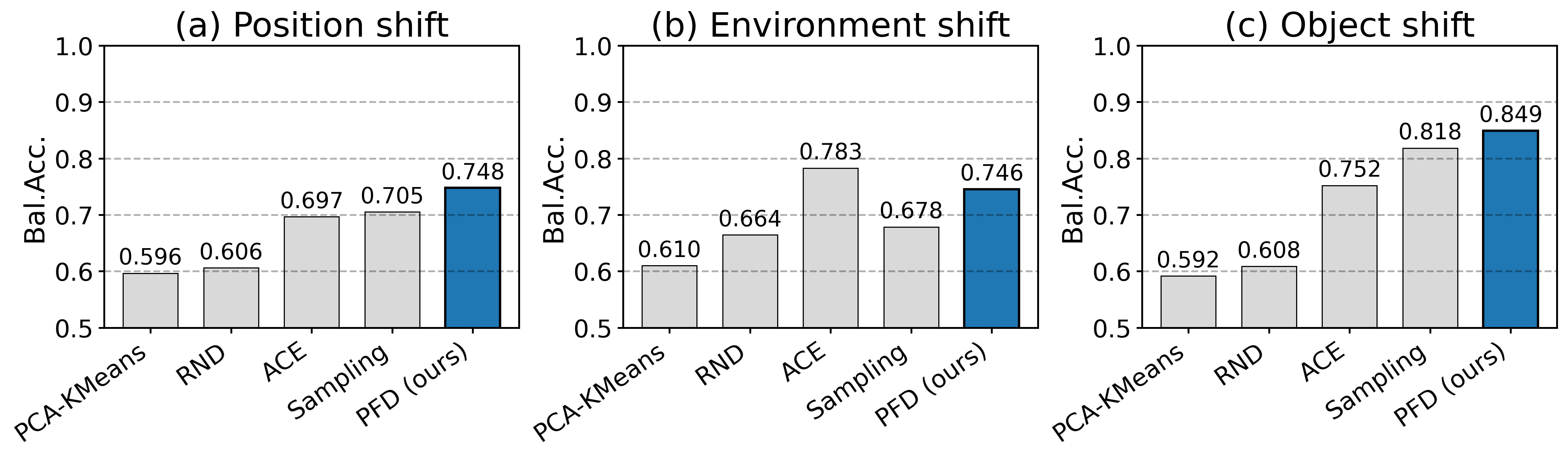}\\[-1mm]
        \small (b) Balanced Accuracy
    \end{minipage}

    \caption{
    Failure detection performance averaged across LIBERO-PRO suites
    within each distribution-shift category.
    \method{} is highlighted in blue, while baseline methods are shown in gray.
    }
    \label{fig:performance_summary}
    \vspace{-2mm}
\end{figure*}

A common and practical way to estimate epistemic uncertainty is through deep ensembles, which measure disagreement across multiple independently trained model instances~\cite{lakshminarayanan2017deep}.
However, their computational and storage costs can become prohibitive when scaling to large pretrained foundation models~\cite{papamarkou2024bayesian}. 
Recent work has extended this idea to flow-based VLAs by separately fine-tuning multiple VLA models and measuring disagreement in their predicted flow velocities~\cite{romer2026uqvla}. However, such ensemble-based approaches require training and maintaining multiple sets of VLA parameters.
As a more efficient alternative, post-hoc perturbation methods have been explored for epistemic uncertainty estimation, particularly in large language models~\cite{wang2024epistemic,liu2026noise,nie2025epistemic,shi2025tfb,zhang2026tokur}. Their application to continuous-action VLA models, however, remains largely unexplored.

In this work, we propose perturbation-based failure detection
(\method), a training-free framework for epistemic uncertainty estimation in pretrained VLA models.
Rather than relying on stochastic action-generation
variability, \method{} constructs local model variations around the pretrained VLA parameters through low-rank perturbations of selected weight matrices.
As shown in Fig.~\ref{fig:method}, disagreement among the resulting perturbed action predictions is used as an epistemic uncertainty signal for failure detection.
Experiments on LIBERO~\cite{libero} and
LIBERO-PRO~\cite{zhou2025liberopro} show that \method{} achieves the
highest average AUROC and balanced accuracy among the evaluated methods while consistently outperforming stochastic action sampling
across distribution shifts.
Real-world robot experiments further demonstrate the competitive performance of \method{} under an unseen object shift.

\section{Related Work}
\subsection{Epistemic Uncertainty Quantification}

Epistemic uncertainty reflects uncertainty over model parameters or model knowledge and is commonly studied through Bayesian neural networks and variational inference~\cite{blundell2015weight,graves2011practical,kendall2017uncertainties}.
In practice, Monte Carlo dropout and deep ensembles are widely used to estimate epistemic uncertainty.
Monte Carlo dropout estimates uncertainty through repeated dropout-enabled inference at test time~\cite{gal2016dropout}, which limits its direct application to pretrained VLA models that were not trained with dropout.
Deep ensembles estimate epistemic uncertainty from disagreement across independently trained models~\cite{lakshminarayanan2017deep,papamarkou2024bayesian}, but maintaining and evaluating multiple models can be computationally expensive for large pretrained foundation models.
To address these limitations, post-hoc perturbation methods have been explored as an efficient way to quantify uncertainty without retraining or maintaining multiple pretrained models~\cite{liu2026noise,wang2024epistemic,nie2025epistemic}.
Recent work has also used low-rank parameterizations for scalable uncertainty estimation, including rank-1 Bayesian neural networks~\cite{dusenberry2020rank1}, Bayesian low-rank adaptation (LoRA)~\cite{wang2024blob}, training-free Bayesianization of pretrained LoRA adapters~\cite{shi2025tfb}, Bayesian low-rank parameterizations~\cite{toure2026singular}, and inference-time weight perturbations within a low-rank subspace~\cite{zhang2026tokur}.
Although these approaches have been explored in neural networks and large pretrained models, their application to continuous-action VLA policies has received limited attention.

\subsection{Failure Detection in VLA Models}

Reliable deployment of VLA models requires detecting
failure-relevant signals during policy execution.
Prior approaches such as SAFE~\cite{gu2025safe}
and recent VLA calibration methods~\cite{francis2026temporal}
rely on labeled success and failure trajectories.
However, collecting sufficiently large and diverse sets of failure trajectories can be costly and potentially unsafe, and such data may not cover failures that arise under previously unseen deployment conditions.
Recent work has explored failure detection through
representation-based out-of-distribution (OOD) detection,
action uncertainty, and temporal consistency.
OOD detection approaches include PCA-KMeans,
which measures distances from projected embeddings to
clusters fitted on in-distribution data
~\cite{liu2024siriusfleet}, and Random Network Distillation
(RND) for detecting unfamiliar representations
~\cite{roemer2025fiper}.
FIPER~\cite{roemer2025fiper} quantifies action uncertainty
using Action Conditional Entropy (ACE) over stochastic
action samples generated by generative policies.
Sentinel~\cite{agia2024sentinel} monitors generative robot
policies by combining temporal action consistency with
VLM-based task progress assessment.
Other approaches leverage token-level entropy or perplexity
from autoregressive action distributions
~\cite{karli2025ask}.
In contrast, \method{} uses action disagreement induced by low-rank perturbations of pretrained VLA parameters as an epistemic uncertainty signal for failure detection.

\section{Background}

\subsection{Vision-Language-Action Policy}

We consider a Vision-Language-Action policy $\pi_\theta$
parameterized by $\theta$.
At timestep $t$, the policy receives a multimodal observation
\begin{equation}
x_t = (v_t, s_t, \ell),
\end{equation}
where $v_t$ denotes visual observations, $s_t$ denotes the
robot proprioceptive state, and $\ell$ denotes the language
instruction.

Given $x_t$, the policy predicts a chunk of future continuous
actions:
\begin{equation}
\mathbf{a}_{t:t+H-1}
=
\pi_\theta(x_t;\mathbf z),
\end{equation}
where $\mathbf z$ denotes the action-generation noise for
stochastic generative policies and is omitted for deterministic
policies. Here, $H$ denotes the action chunk horizon, $d$ denotes
the action dimension, and $\mathbf{a}_t \in \mathbb{R}^{d}$ denotes
a continuous robot action at timestep $t$.

\subsection{Bayesian Uncertainty Decomposition}

From a Bayesian perspective, model parameters
$\theta \sim p(\theta \mid \mathcal D)$ are treated as random
variables conditioned on training data
$\mathcal D$.
Predictive uncertainty can then be decomposed into
aleatoric uncertainty and epistemic uncertainty
under the posterior distribution over model
parameters~\cite{kendall2017uncertainties,gal2016dropout}:

\begin{equation}
\begin{aligned}
\mathrm{Var}(\mathbf a \mid x_t,\mathcal D)
&=
\underbrace{
\mathbb E_{\theta \sim p(\theta \mid \mathcal D)}
\left[
\mathrm{Var}(\mathbf a \mid x_t,\theta)
\right]
}_{\mathrm{Aleatoric}}
\\
&\quad+
\underbrace{
\mathrm{Var}_{\theta \sim p(\theta \mid \mathcal D)}
\left[
\mathbb E(\mathbf a \mid x_t,\theta)
\right]
}_{\mathrm{Epistemic}}.
\end{aligned}
\end{equation}

The epistemic term reflects uncertainty over plausible
model parameters under the posterior distribution.
By Bayes' rule, the parameter posterior can be expressed as
\begin{equation}
p(\theta \mid \mathcal D)
=
\frac{p(\mathcal D \mid \theta)p(\theta)}
{p(\mathcal D)},
\qquad
p(\mathcal D)
=
\int p(\mathcal D \mid \theta)p(\theta)\,d\theta .
\end{equation}
For large-scale pretrained foundation models, evaluating
the marginal likelihood $p(\mathcal D)$ requires integration
over a high-dimensional parameter space, making exact
posterior inference computationally intractable
~\cite{papamarkou2024bayesian,liu2026noise,
wang2024epistemic,nie2025epistemic}.
Therefore, we focus on training-free approximations of
local model variations without retraining the
pretrained VLA model.

\section{Methodology}

\subsection{Low-Rank Parameter Perturbation}

Following prior work on perturbation-based uncertainty
estimation~\cite{wang2024epistemic,liu2026noise,zhang2026tokur},
we approximate local parameter variations around the pretrained
parameters by directly perturbing selected weight matrices of
the VLA policy.
For a pretrained weight matrix
$W_0 \in \mathbb{R}^{d_{\mathrm{out}}\times d_{\mathrm{in}}}$,
a full weight matrix perturbation requires sampling and storing
$O(d_{\mathrm{out}}d_{\mathrm{in}})$ additional perturbation
parameters for each model variant.
This becomes computationally expensive for large projection
matrices in pretrained foundation models.
We therefore construct each perturbation using a rank-$r$
factorization, where
$r\ll\min(d_{\mathrm{out}},d_{\mathrm{in}})$.

The $k$-th low-rank weight perturbation is defined as
\begin{equation}
\Delta W^{(k)}
=
\frac{\sigma}{\sqrt{d_{\mathrm{in}}}}
A^{(k)}
\left(B^{(k)}\right)^\top,
\label{eq:low_rank_perturbation}
\end{equation}
where
$A^{(k)}\in\mathbb{R}^{d_{\mathrm{out}}\times r}$,
$B^{(k)}\in\mathbb{R}^{d_{\mathrm{in}}\times r}$,
$\sigma$ controls the perturbation strength,
$1/\sqrt{d_{\mathrm{in}}}$ prevents the perturbation scale from growing with $d_{\mathrm{in}}$,
and the entries of $A^{(k)}$ and $B^{(k)}$ are sampled i.i.d. from a
standard Gaussian distribution \(\mathcal N(0,1)\).

The resulting perturbed weight matrix is
\begin{equation}
W^{(k)}
=
W_0+\Delta W^{(k)}.
\end{equation}

Since $\Delta W^{(k)}$ is parameterized by rank-$r$ factors,
it satisfies
$\mathrm{rank}(\Delta W^{(k)})\leq r$.
The low-rank factorization reduces the perturbation
parameterization from
$O(d_{\mathrm{out}}d_{\mathrm{in}})$ to
$O(r(d_{\mathrm{out}}+d_{\mathrm{in}}))$,
enabling efficient generation of multiple local parameter
variations.
Each perturbation-specific factor pair
$A^{(k)}$ and $B^{(k)}$ is sampled once and kept fixed
throughout the episode rather than being resampled at every
timestep.
This allows each perturbation sample to represent a temporally
consistent local model variation.

\subsection{Perturbation-Based Uncertainty Estimation}

Applying low-rank perturbations to the selected weight matrices
produces a perturbed parameter set $\theta^{(k)}$.
We construct $K$ model variants consisting of the unperturbed
base policy and $K-1$ perturbed variants.
For stochastic generative policies, we fix the generation
noise $\mathbf z_0$ across model variants to isolate the effect of
parameter variation and obtain
\begin{equation}
\mathbf a^{(k)}_{t:t+H-1}
=
\pi_{\theta^{(k)}}(x_t;\mathbf z_0).
\end{equation}

We then quantify uncertainty as disagreement across perturbed
action chunks:
\begin{equation}
u_t
=
\frac{1}{Hd}
\sum_{h=0}^{H-1}
\sum_{j=1}^{d}
\mathrm{Var}_{k}
\left[
a_{t+h,j}^{(k)}
\right].
\label{eq:pfd_uncertainty}
\end{equation}

Here, $a_{t+h,j}^{(k)}$ denotes the $j$-th action dimension
of the predicted action at future step $h$ under model variant $k$.
Variance is computed across model variants and averaged over
chunk steps and action dimensions.
The perturbed variants are used only for uncertainty estimation,
while actions predicted by the unperturbed base policy are executed
in the environment.

\textbf{Intuition.} Local parameter variations can reveal how strongly the policy is constrained by the training data. When the policy behavior for a given observation is well constrained, nearby model variations are expected to produce consistent action predictions. 
In contrast, observations that are insufficiently constrained during training may lead to greater disagreement across perturbed models. Such disagreement can therefore provide a useful epistemic signal for detecting failure-prone situations.

\subsection{Temporal Aggregation for Failure Detection}

To obtain a stable uncertainty estimate for the current
timestep, we aggregate recent chunk-level uncertainty
scores using a temporal sliding
window~\cite{roemer2025fiper}:
\begin{equation}
U_t
=
\frac{1}{T_w}
\sum_{\tau=t-T_w+1}^{t}
u_{\tau},
\label{eq:temporal}
\end{equation}
where $T_w$ denotes the sliding window size.
This temporal aggregation captures persistent
disagreement patterns rather than transient action
fluctuations during policy execution. 
We refer to $U_t$ as the timestep-level failure score used
for online failure detection.

\section{Experiments}

\subsection{Models and Perturbation Design}

\textbf{Models.}
We evaluate $\pi_{0.5}$~\cite{black2025pi05} in the main
experiments and OpenVLA-OFT~\cite{kim2025oft} in the
perturbation-strength analysis.
OpenVLA-OFT predicts continuous action chunks through a
deterministic regression head, whereas $\pi_{0.5}$ uses a
flow-matching-based generative policy that supports stochastic
action sampling.
In simulation, $\pi_{0.5}$ predicts action chunks of length 10,
while OpenVLA-OFT predicts action chunks of length 8.
For both policies, all predicted actions in each chunk are executed
before the next policy inference.

\textbf{Perturbation Location.}
We apply low-rank weight perturbations to the MLP down-projection in all transformer layers, leaving the first layer unperturbed to preserve the initial feature representation.
These weight matrices constitute the perturbation set
$\mathcal{S}$ used throughout our method.
In OpenVLA-OFT, the perturbations are applied to the dual vision
encoder, whereas in $\pi_{0.5}$, they are applied to the VLM expert
before the action expert module.
This design targets uncertainty in the model's understanding of the current
observation and its effects on downstream action predictions.

\textbf{Implementation Details.}
We use a perturbation rank of $r=64$, which is substantially smaller
than the maximum rank
$\min(d_{\mathrm{out}},d_{\mathrm{in}})=2048$
of the perturbed MLP projection matrices in $\pi_{0.5}$.
For $\pi_{0.5}$, the perturbation strength is set to $\sigma=0.6$
and fixed across all main experiments.
Unless otherwise specified, we use $K=5$ model variants, consisting
of the unperturbed base model and four low-rank perturbed variants.
All experiments are conducted using a single NVIDIA GeForce RTX 3090 GPU.
We analyze the sensitivity to $\sigma$ in
Sec.~\ref{sec:perturbation_ablation}.

\begin{table*}[t]
\centering
\caption{
Failure detection performance under LIBERO-PRO distribution shifts ($K=5$).
Detection time is averaged over detected failures only.
Detection-time values are parenthesized when either TPR or TNR is below $0.4$ and are excluded from the average detection time.
The best and second-best valid results are shown in bold and
underlined, respectively.
Higher is better for AUROC and balanced accuracy; lower is better for detection time.
}
\label{tab:main_results}

\footnotesize
\renewcommand{\arraystretch}{0.1}
\setlength{\aboverulesep}{0.1ex}
\setlength{\belowrulesep}{0.1ex}
\setlength{\tabcolsep}{4pt}

\begin{tabular*}{\textwidth}{
@{\extracolsep{\fill}}
llccccc
@{}
}
\toprule
\textbf{Environment} & \textbf{Metric}
& \textbf{PCA-KMeans}
& \textbf{RND}
& \textbf{ACE}
& \textbf{Sampling}
& \textbf{\method~(Ours)} \\
\midrule

\multicolumn{7}{c}{\textit{Position Shift}} \\
\midrule

\multirow{3}{*}{\shortstack[l]{LIBERO-PRO Spatial-Pos\\Success Rate: 43.8\%}}
& AUROC $\uparrow$
& 0.782
& 0.815
& 0.837
& \underline{0.943}
& \textbf{0.961} \\
& Bal. Acc. $\uparrow$
& 0.672
& 0.657
& 0.628
& \underline{0.883}
& \textbf{0.902} \\
& Det. Time $\downarrow$
& 74.89
& 74.03
& (73.45)
& \textbf{63.17}
& \underline{69.39} \\
\midrule

\multirow{3}{*}{\shortstack[l]{LIBERO-PRO Object-Pos\\Success Rate: 21.0\%}}
& AUROC $\uparrow$
& 0.369
& \underline{0.604}
& 0.428
& 0.484
& \textbf{0.640} \\
& Bal. Acc. $\uparrow$
& 0.495
& 0.503
& 0.499
& \underline{0.509}
& \textbf{0.613} \\
& Det. Time $\downarrow$
& (68.83)
& (69.64)
& (51.22)
& (41.58)
& \textbf{53.02} \\
\midrule

\multirow{3}{*}{\shortstack[l]{LIBERO-PRO Goal-Pos\\Success Rate: 32.8\%}}
& AUROC $\uparrow$
& 0.648
& 0.859
& \underline{0.885}
& 0.848
& \textbf{0.891} \\
& Bal. Acc. $\uparrow$
& 0.522
& 0.757
& \textbf{0.814}
& 0.740
& \underline{0.793} \\
& Det. Time $\downarrow$
& (65.56)
& 56.28
& \underline{50.92}
& \textbf{37.97}
& 53.48 \\
\midrule

\multirow{3}{*}{\shortstack[l]{LIBERO-PRO 10-Pos\\Success Rate: 11.0\%}}
& AUROC $\uparrow$
& 0.583
& 0.614
& \textbf{0.911}
& 0.819
& \underline{0.852} \\
& Bal. Acc. $\uparrow$
& 0.535
& 0.575
& \textbf{0.793}
& \underline{0.668}
& 0.665 \\
& Det. Time $\downarrow$
& (45.09)
& (73.12)
& 63.16
& \underline{59.83}
& \textbf{56.88} \\
\midrule

\multicolumn{7}{c}{\textit{Environment Shift}} \\
\midrule

\multirow{3}{*}{\shortstack[l]{LIBERO-PRO Spatial-Env\\Success Rate: 42.2\%}}
& AUROC $\uparrow$
& \textbf{0.948}
& \underline{0.927}
& 0.888
& 0.794
& 0.823 \\
& Bal. Acc. $\uparrow$
& 0.682
& \underline{0.760}
& \textbf{0.772}
& 0.649
& 0.727 \\
& Det. Time $\downarrow$
& (70.37)
& 69.55
& \underline{61.07}
& (35.37)
& \textbf{46.42} \\
\midrule

\multirow{3}{*}{\shortstack[l]{LIBERO-PRO Object-Env\\Success Rate: 75.8\%}}
& AUROC $\uparrow$
& 0.409
& 0.559
& \underline{0.779}
& 0.775
& \textbf{0.803} \\
& Bal. Acc. $\uparrow$
& 0.508
& 0.508
& 0.608
& \textbf{0.681}
& \underline{0.667} \\
& Det. Time $\downarrow$
& (3.57)
& (3.57)
& (48.09)
& \textbf{41.28}
& \underline{42.64} \\
\midrule

\multirow{3}{*}{\shortstack[l]{LIBERO-PRO Goal-Env\\Success Rate: 51.6\%}}
& AUROC $\uparrow$
& 0.776
& 0.832
& \textbf{0.906}
& 0.736
& \underline{0.874} \\
& Bal. Acc. $\uparrow$
& 0.531
& 0.669
& \textbf{0.833}
& 0.603
& \underline{0.798} \\
& Det. Time $\downarrow$
& (71.88)
& (47.70)
& \underline{41.59}
& \textbf{35.30}
& 41.95 \\
\midrule

\multirow{3}{*}{\shortstack[l]{LIBERO-PRO 10-Env\\Success Rate: 70.8\%}}
& AUROC $\uparrow$
& 0.624
& 0.747
& \textbf{0.921}
& 0.838
& \underline{0.872} \\
& Bal. Acc. $\uparrow$
& 0.569
& 0.665
& \textbf{0.770}
& 0.686
& \underline{0.713} \\
& Det. Time $\downarrow$
& (73.65)
& (70.75)
& 66.61
& \textbf{55.99}
& \underline{57.29} \\
\midrule

\multicolumn{7}{c}{\textit{Object Shift}} \\
\midrule

\multirow{3}{*}{\shortstack[l]{LIBERO-PRO Spatial-Obj\\Success Rate: 97.8\%}}
& AUROC $\uparrow$
& 0.873
& \textbf{0.968}
& 0.844
& 0.877
& \underline{0.947} \\
& Bal. Acc. $\uparrow$
& 0.810
& \underline{0.811}
& 0.631
& 0.806
& \textbf{0.905} \\
& Det. Time $\downarrow$
& \underline{68.83}
& 70.13
& (63.64)
& \textbf{44.89}
& 71.21 \\
\midrule

\multirow{3}{*}{\shortstack[l]{LIBERO-PRO Object-Obj\\Success Rate: 93.6\%}}
& AUROC $\uparrow$
& 0.559
& 0.630
& 0.848
& \underline{0.948}
& \textbf{0.958} \\
& Bal. Acc. $\uparrow$
& 0.525
& 0.515
& 0.537
& \underline{0.784}
& \textbf{0.893} \\
& Det. Time $\downarrow$
& (3.57)
& (28.57)
& (89.29)
& \underline{64.11}
& \textbf{54.50} \\
\midrule

\multirow{3}{*}{\shortstack[l]{LIBERO-PRO Goal-Obj\\Success Rate: 84.8\%}}
& AUROC $\uparrow$
& 0.561
& 0.732
& \textbf{0.951}
& 0.856
& \underline{0.892} \\
& Bal. Acc. $\uparrow$
& 0.478
& 0.599
& \textbf{0.848}
& \underline{0.796}
& 0.750 \\
& Det. Time $\downarrow$
& (--)
& (66.14)
& \underline{51.36}
& \textbf{35.93}
& 62.59 \\
\midrule

\multirow{3}{*}{\shortstack[l]{LIBERO-PRO 10-Obj\\Success Rate: 66.6\%}}
& AUROC $\uparrow$
& 0.362
& 0.542
& \textbf{0.914}
& 0.849
& \underline{0.878} \\
& Bal. Acc. $\uparrow$
& 0.492
& 0.545
& \textbf{0.740}
& \underline{0.738}
& 0.689 \\
& Det. Time $\downarrow$
& (75.44)
& (77.15)
& 62.93
& \textbf{54.92}
& \underline{58.42} \\
\midrule
\midrule

\multirow{3}{*}{\textbf{Average}}
& AUROC $\uparrow$
& 0.625
& 0.736
& \underline{0.843}
& 0.814
& \textbf{0.866} \\
& Bal. Acc. $\uparrow$
& 0.568
& 0.630
& 0.706
& \underline{0.712}
& \textbf{0.760} \\
& Det. Time $\downarrow$
& 71.86
& 67.50
& 56.81
& \textbf{49.34}
& \underline{55.65} \\

\bottomrule
\vspace{-2mm}
\end{tabular*}
\end{table*}

\subsection{Evaluation Benchmarks}

\textbf{LIBERO.}
LIBERO~\cite{libero} is a language-conditioned robot manipulation benchmark comprising diverse manipulation tasks.
We evaluate four commonly used suites: \textit{Spatial}, \textit{Object}, \textit{Goal}, and \textit{LIBERO-10}, which emphasize spatial relationships, object interactions, task goals, and diverse long-horizon manipulation tasks, respectively.

\textbf{LIBERO-PRO.}
LIBERO-PRO~\cite{zhou2025liberopro} extends LIBERO with controlled distribution shifts while preserving task semantics.
We consider three shift categories: \textit{Position} shifts, which modify object placements and relative spatial arrangements; \textit{Environment} shifts, which alter the visual scene context without changing the underlying task; and \textit{Object} shifts, which modify non-essential object attributes such as color, texture, or size while preserving semantic identity.
These shifts frequently induce failures in pretrained VLA policies despite their strong performance on standard LIBERO, making LIBERO-PRO a challenging benchmark for evaluating failure detection under distribution shift.

\subsection{Failure Detection Baselines}

We compare \method{} against four baselines: two representation-based OOD detection methods,
PCA-KMeans and Random Network Distillation (RND), and two action-uncertainty methods, Action Conditional Entropy (ACE) and stochastic action sampling.
These baselines cover both representation-level novelty detection and action-level uncertainty, with stochastic sampling serving as the most direct counterpart to our approach.
For fair comparison, all baseline scores are processed using the same temporal aggregation and failure-detection protocol as our method.

\textbf{PCA-KMeans.}
We adopt the PCA-KMeans baseline introduced in ~\cite{liu2024siriusfleet}, following its use in
SAFE~\cite{gu2025safe}.
Let $e_t$ denote the final-layer action-token embedding averaged over
the predicted action chunk.
Using successful in-distribution reference embeddings, we learn a PCA
projection $P$ and fit K-means centroids $\{c_j\}$.
At test time, the OOD score is the distance to the nearest centroid,
$u_t^{\mathrm{PCA}}=\min_j\|Pe_t-c_j\|_2$.

\textbf{RND.}
We adapt the RND-OE implementation from FIPER~\cite{roemer2025fiper}
using the same embeddings $e_t$.
The predictor is trained on successful in-distribution embeddings to match a fixed randomly initialized target network.
The OOD score is the prediction error,
$u_t^{\mathrm{RND}}=
\|f_{\mathrm{pred}}(e_t)-f_{\mathrm{target}}(e_t)\|_2$.

\textbf{Sampling.}
For each observation, we generate $K$ stochastic action chunks by
independently sampling the initial flow-matching noise,
$\mathbf{a}^{(k)}_{t:t+H-1}
=\pi_{\theta}(x_t;\mathbf{z}^{(k)})$ with
$\mathbf{z}^{(k)}\sim\mathcal{N}(\mathbf{0},\mathbf{I})$.
For direct comparison, we compute the same variance-based uncertainty
score as our method over all action dimensions
$(x,y,z,r_x,r_y,r_z,g)$.
By construction, sampling varies action-generation noise under fixed
model parameters, whereas \method{} varies model parameters with fixed
generation noise.

\textbf{ACE.}
We implement ACE, originally introduced in FIPER~\cite{roemer2025fiper},
following the implementation of~\cite{romer2026uqvla}.
Using the same $K$ stochastic action chunks, ACE obtains end-effector
position trajectories $(x,y,z)$ directly from Cartesian actions or
through forward kinematics for joint-space actions.
ACE then discretizes the 3D positions and averages the resulting
spatial entropy over the prediction horizon.

\subsection{Evaluation Protocol}
\label{sec:evaluation_protocol}

\textbf{Evaluation Rollouts.}
For each LIBERO-PRO suite, we evaluate 500 rollouts across 10 tasks, yielding 6,000 evaluation rollouts over the 12 distribution-shift suites.

\textbf{Reference and Calibration Sets.}
For each LIBERO suite, we construct separate reference and calibration
sets from successful in-distribution rollouts and use them for evaluating
the corresponding LIBERO-PRO shifts.
For example, for the LIBERO-Spatial suite, we collect
\textbf{Reference set}: 5 successful rollouts per task
(50 rollouts total); and
\textbf{Calibration set}: another 5 successful rollouts per task
(50 rollouts total) for conformal threshold calibration.
The same LIBERO-Spatial reference and calibration sets are used for
evaluating Spatial-Pos, Spatial-Env, and Spatial-Obj.
The reference and calibration sets consist of disjoint rollouts.

\textbf{Reference Set Usage.}
The reference set is used to fit PCA-KMeans and RND and to calibrate
the spatial discretization used by ACE.
For PCA-KMeans and RND, we use the same chunk-averaged
final-layer action-token embeddings to fit the corresponding OOD detector.
In contrast, \method{} does not use the reference set and requires
no auxiliary model fitting.

\textbf{Failure Score and Conformal Calibration.}
For each method, let $u_{n,t}$ denote the raw uncertainty score
at timestep $t$ of rollout $n$.
We temporally aggregate $u_{n,t}$ according to
Eq.~\eqref{eq:temporal} using a sliding window of size $T_w=5$,
producing the timestep-level failure score $U_{n,t}$.
For threshold calibration and AUROC computation, we define the
rollout-level score as the maximum failure score observed during
execution, $U_n^{\max}=\max_t U_{n,t}$.
We treat $U_n^{\max}$ as a rollout-level nonconformity score and
calibrate a constant threshold from successful in-distribution
rollouts using split conformal prediction
~\cite{angelopoulos2023conformal}.
Given $N_{\mathrm{cal}}$ calibration rollouts and significance level
$\alpha=0.05$, we compute the threshold using the conformal $95\%$ quantile:
\begin{equation}
k =
\left\lceil
(N_{\mathrm{cal}}+1)(1-\alpha)
\right\rceil,
\qquad
\delta = U_{(k)}^{\max},
\end{equation}
where $U_{(k)}^{\max}$ denotes the $k$-th smallest rollout-level score
in the calibration set.
Under exchangeability, the probability of raising a false failure alarm during a successful in-distribution rollout is at most $\alpha$~\cite{angelopoulos2023conformal}.
We use a constant threshold because distribution shifts can alter
execution speed and task progress, making timestep-wise alignment
between calibration and evaluation rollouts unreliable.
The threshold is transferred unchanged to the corresponding
LIBERO-PRO shifts without using any LIBERO-PRO rollouts for calibration.
However, the conformal guarantee applies only to successful
in-distribution rollouts; performance under distribution shift is evaluated empirically.

\begin{figure*}[t]
    \centering

    \begin{minipage}[t]{0.49\textwidth}
        \centering
        \includegraphics[width=\linewidth]
        {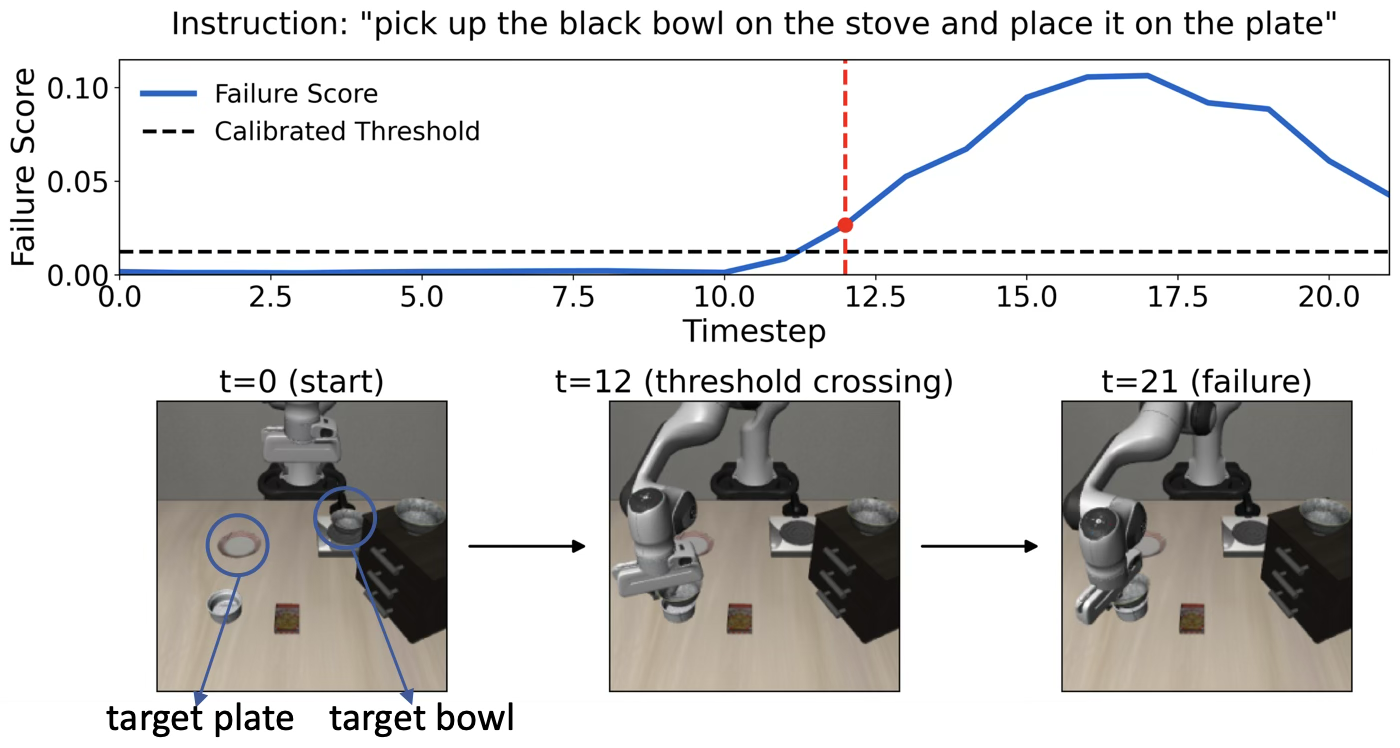}
        \vspace{-2mm}

        {\small \textbf{(a)} Failure case: LIBERO-PRO Spatial-Pos}
    \end{minipage}
    \hfill
    \begin{minipage}[t]{0.49\textwidth}
        \centering
        \includegraphics[width=\linewidth]
        {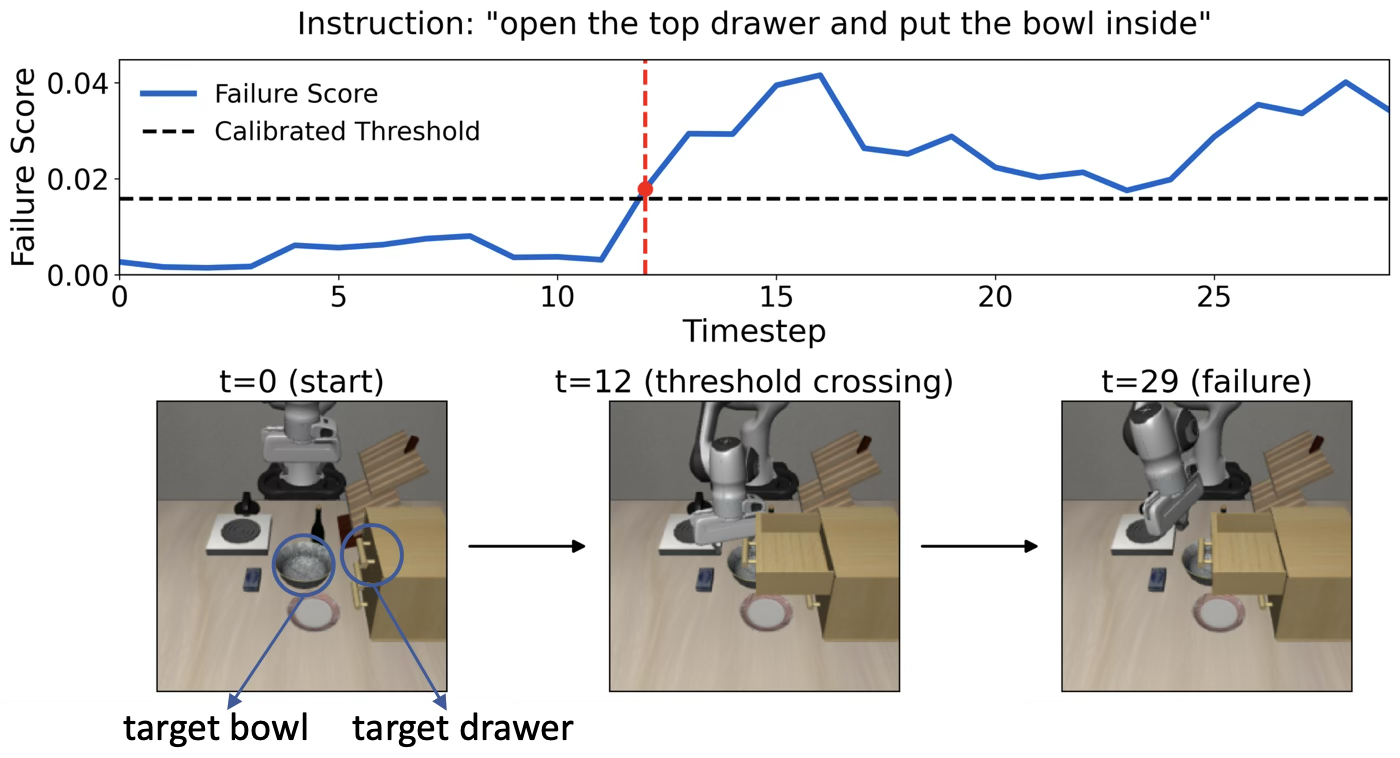}
        \vspace{-2mm}

        {\small \textbf{(b)} Failure case: LIBERO-PRO Goal-Obj}
    \end{minipage}
    \caption{
    Representative failure cases from LIBERO-PRO. Blue curves show the \method{} failure scores, black dashed lines denote split-conformal thresholds ($\alpha=0.05$), and red dashed lines indicate the first threshold crossing. Images show the start, first threshold crossing, and final failure timestep.
    }
    \label{fig:qualitative_cases}
    \vspace{-2mm}
\end{figure*}

\begin{figure}[t]
    \centering

    \begin{minipage}{0.49\linewidth}
        \centering
        \includegraphics[width=\linewidth]{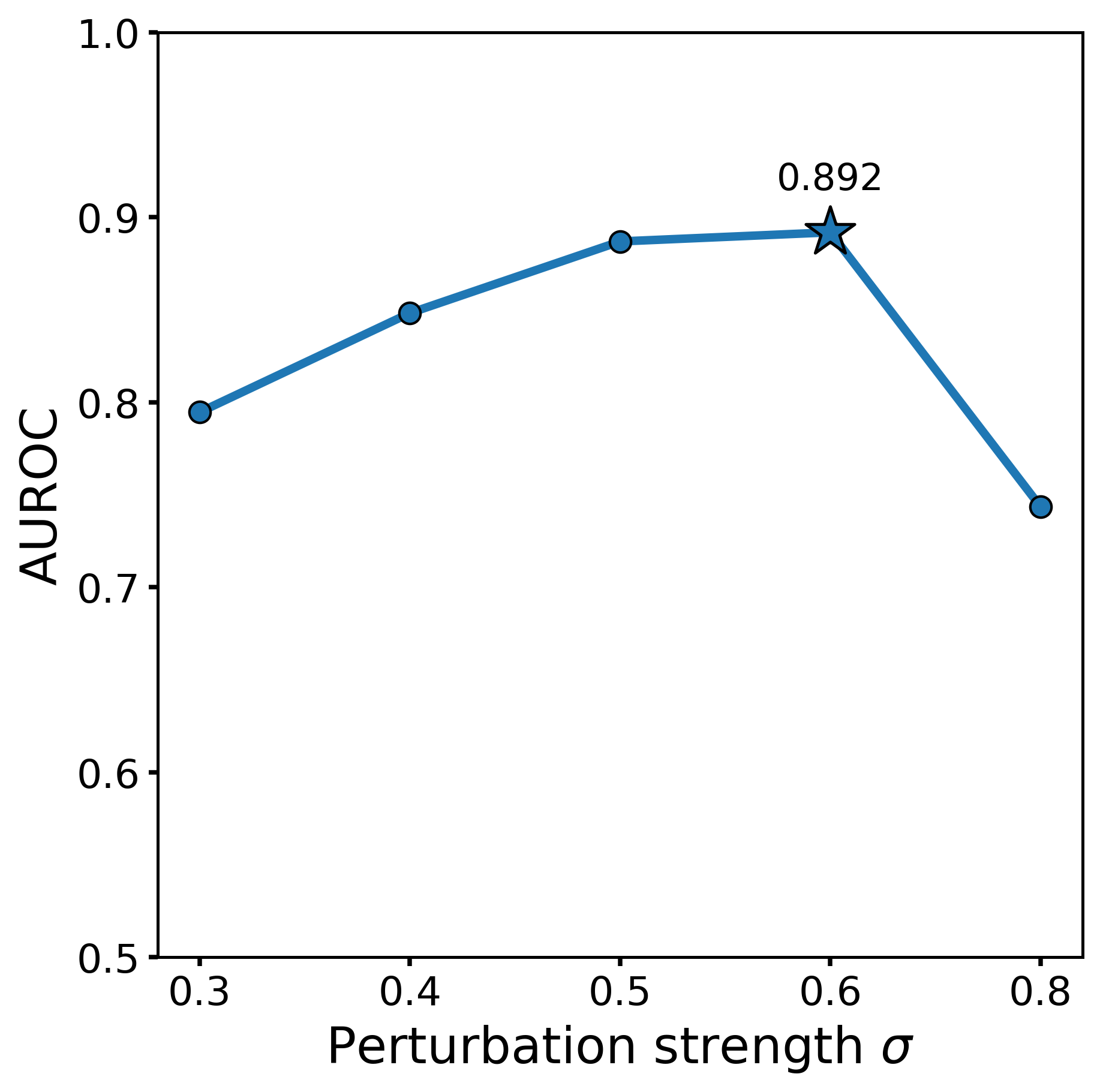}

        \vspace{-1mm}
        {\small (a) $\pi_{0.5}$}
    \end{minipage}
    \hfill
    \begin{minipage}{0.49\linewidth}
        \centering
        \includegraphics[width=\linewidth]{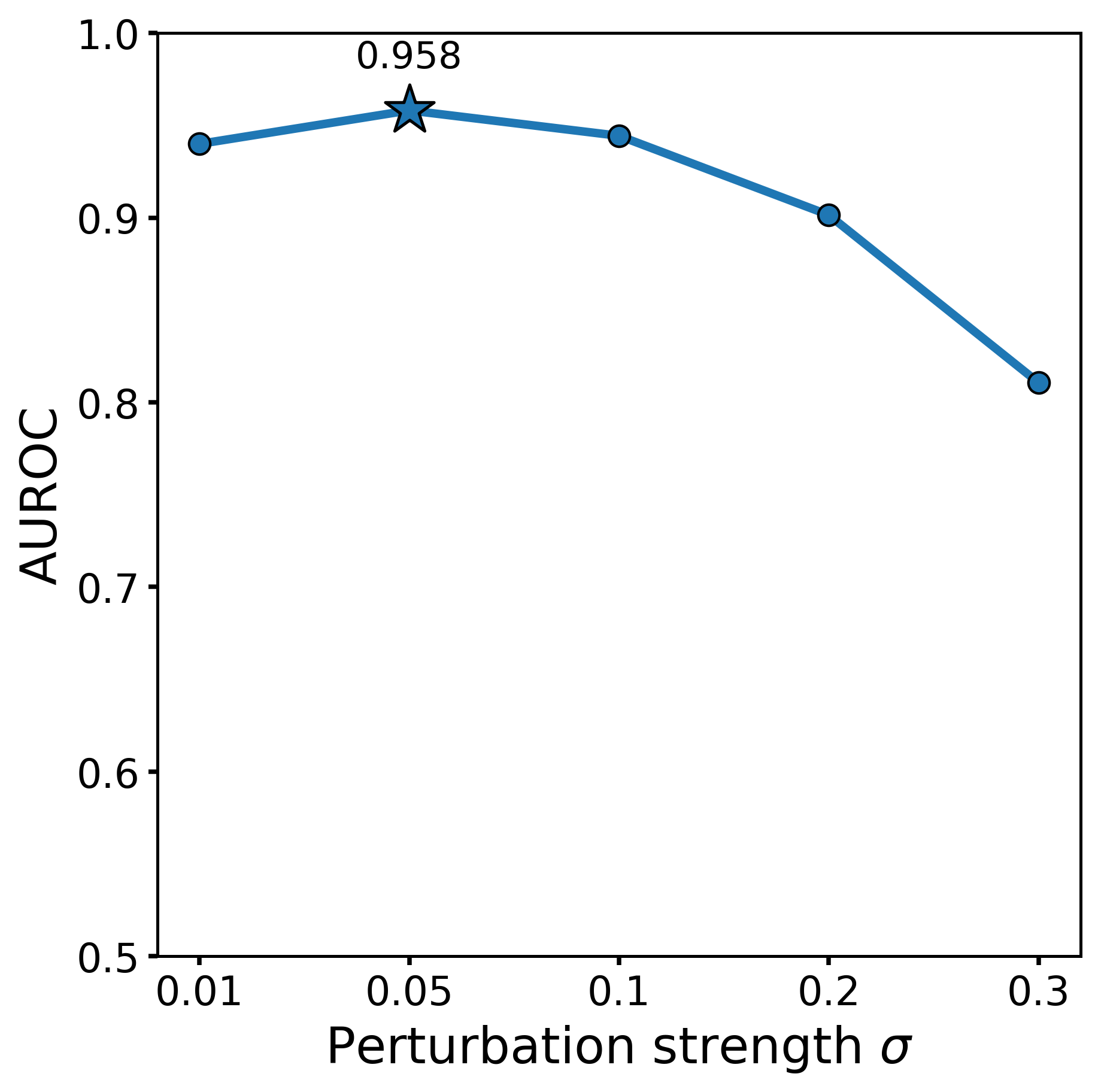}

        \vspace{-1mm}
        {\small (b) OpenVLA-OFT}
    \end{minipage}

    \vspace{-1mm}
    \caption{
    Sensitivity of failure-detection AUROC to perturbation strength $\sigma$
    on LIBERO-10 ($K=2$).
    Stars indicate the best-performing strength for each model.
    }
    \label{fig:sigma_sweep}
    \vspace{-2mm}
\end{figure}

\subsection{Evaluation Metrics}

We evaluate failure detection using three metrics.

\textbf{AUROC.}
AUROC is a threshold-free metric that evaluates the overall
discrimination between successful and failed rollouts.
Given rollout-level scores $U_n^{\max}$ and failure labels
$y_n\in\{0,1\}$, it measures the ability of the scores to separate
the two outcomes across all possible thresholds.
Higher values indicate better discrimination.

\textbf{Balanced Accuracy.}
Balanced accuracy evaluates threshold-based runtime failure
detection performance.
Rollout $n$ is classified as a failure if $U_{n,t}>\delta$ at any
timestep during execution.
Balanced accuracy is defined as
\begin{equation}
\mathrm{Bal.\ Acc.}
=
\frac{1}{2}
\left(
\mathrm{TPR}+\mathrm{TNR}
\right).
\end{equation}
It gives equal weight to the detection rate of failed rollouts (TPR)
and the correct classification rate of successful rollouts (TNR),
accounting for class imbalance.
Higher values indicate better failure detection performance.

\textbf{Detection Time.}
For a detected failed rollout, the normalized detection time is defined as
\begin{equation}
\mathrm{DT}
=
100\cdot\frac{t_{\mathrm{det}}}{T},
\qquad
t_{\mathrm{det}}
=
\min\left\{t\mid U_t>\delta\right\},
\end{equation}
where $T$ denotes the rollout length.
Lower values indicate earlier failure detection.

\subsection{Failure Detection Results}

\textbf{Overall Performance.}
Table~\ref{tab:main_results} shows that \method{} achieves
the highest average AUROC and balanced accuracy among
the evaluated methods, reaching 0.866 and 0.760, respectively.
Compared with stochastic action sampling, its most direct
counterpart, \method{} improves the averages from 0.814 to
0.866 in AUROC and from 0.712 to 0.760 in balanced accuracy,
while outperforming Sampling across all 12 LIBERO-PRO
suites in AUROC. The largest gain occurs on Object-Pos,
from 0.484 to 0.640.

\textbf{Performance across Shift Types.} Fig.~\ref{fig:performance_summary} further summarizes performance across distribution-shift categories. \method{} achieves the best average performance under Position and Object shifts, while ranking second under Environment shifts. This suggests that the proposed uncertainty signal remains competitive across qualitatively different distribution shifts rather than being specialized to a single shift type.

\textbf{Representation-Based OOD Baselines.}
Representation-based OOD baselines exhibit substantial variation
across LIBERO-PRO suites.
This is consistent with representation-level novelty not necessarily
corresponding to task failure under distribution shift, since successful
rollouts may also deviate from the in-distribution reference representations.
For example, RND ranges from 0.968 on Spatial-Obj to 0.542
on 10-Obj, while PCA-KMeans ranges from 0.948 on
Spatial-Env to 0.362 on 10-Obj.

\textbf{Detection Time.}
Finally, \method{} achieves an average normalized detection time of 55.65,
with Sampling detecting failures earlier on average (49.34).
These averages are computed over different valid subsets of suites,
as suites with low TPR or TNR are excluded.

\begin{figure*}[t]
    \centering
    \includegraphics[width=\textwidth]{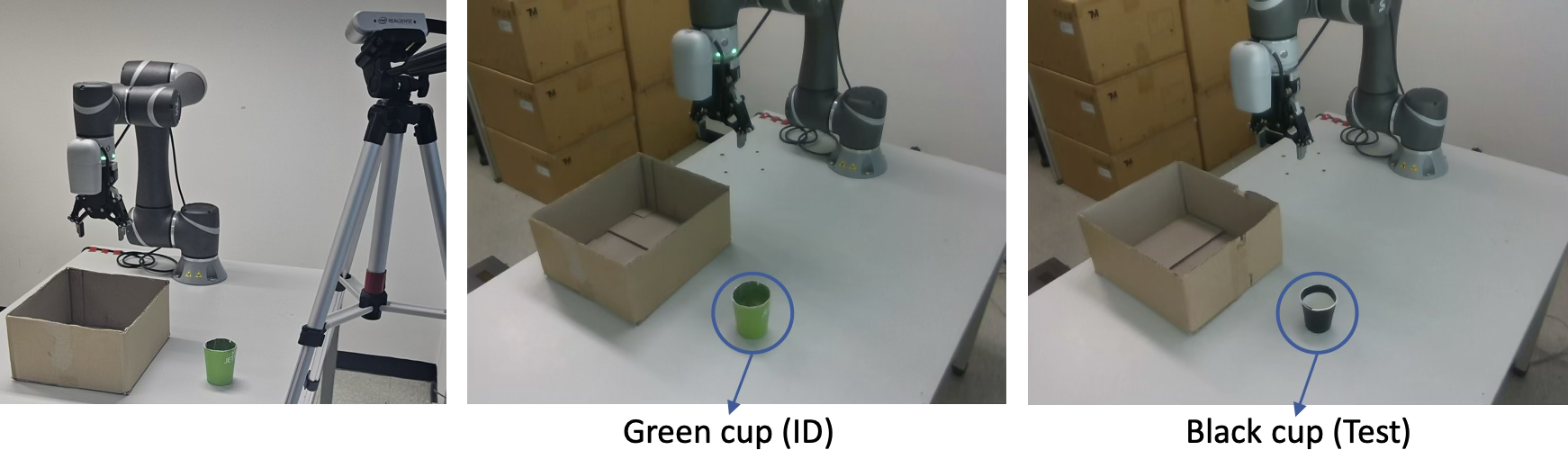}
    \caption{
    Real-world setup and evaluation conditions.
    Left: robot workspace with the TM5-900, AG95 gripper, and    fixed RealSense D435i camera.
    Middle: green-cup ID condition used for reference,   
    calibration, and ablation.
    Right: unseen black-cup condition used for object-shift 
    testing.
    Object positions are randomized within the same workspace 
    region.
    }
    \label{fig:real_robot}
    \vspace{-2mm}
\end{figure*}

\begin{table}[t]
    \centering
    \footnotesize
    \caption{
    Sensitivity to the number of model variants $K$.
    AUROC is averaged across all 12 LIBERO-PRO suites.
    }
    \label{tab:k_ablation}

    \setlength{\tabcolsep}{6pt}
    \renewcommand{\arraystretch}{0.95}

    \begin{tabular}{ccccc}
        \toprule
        \textbf{Method}
        & $K=2$
        & $K=3$
        & $K=4$
        & $K=5$ \\
        \midrule
        Sampling
        & 0.806
        & 0.816
        & 0.813
        & 0.814 \\
        \method{}
        & \textbf{0.852}
        & \textbf{0.860}
        & \textbf{0.863}
        & \textbf{0.866} \\
        \bottomrule
    \end{tabular}
\end{table}

\subsection{Perturbation Parameter Analysis}
\label{sec:perturbation_ablation}

\textbf{Sensitivity to Perturbation Strength.}
We select the perturbation strength $\sigma$ on in-distribution
LIBERO-10 for both models, fixing the selected $\pi_{0.5}$ value
across all 12 LIBERO-PRO suites in the main experiments.
As shown in Fig.~\ref{fig:sigma_sweep}, $\pi_{0.5}$
(success rate: 95.4\%) achieves its highest failure-detection AUROC of
0.892 at $\sigma=0.6$ and remains stable over $\sigma\in[0.5,0.6]$,
whereas OpenVLA-OFT (success rate: 94.0\%) achieves its highest AUROC of
0.958 at $\sigma=0.05$ and remains stable over
$\sigma\in[0.01,0.1]$.
Although the optimal $\sigma$ differs between the two models, both
exhibit a similar sensitivity trend.
This suggests that moderate perturbations provide sufficient model
diversity while remaining close to the pretrained VLA policy.

\textbf{Number of Model Variants.}
We further analyze sensitivity to the number of model variants
$K$ by averaging AUROC across all 12 LIBERO-PRO suites.
As shown in Table~\ref{tab:k_ablation}, \method{} consistently
outperforms stochastic action sampling across all evaluated values of
$K$.
Its average AUROC improves from 0.852 at $K=2$ to 0.866 at $K=5$,
with only marginal gains from additional model variants,
indicating stable performance with a small $K$.

\subsection{Qualitative Visualization}

To complement the quantitative results in
Table~\ref{tab:main_results},
Fig.~\ref{fig:qualitative_cases} shows two representative
LIBERO-PRO failures detected by \method{}.
Under the position shift in Spatial-Pos, the policy successfully grasps
the bowl but places it at the memorized original plate location rather
than the shifted target (Fig.~\ref{fig:qualitative_cases}(a)).
Under the object shift in Goal-Obj, the policy successfully opens the
drawer but fails to grasp the target bowl after its visual attributes are
changed (Fig.~\ref{fig:qualitative_cases}(b)).
In both cases, the \method{} failure score crosses the calibrated
threshold before the task failure occurs.

\begin{figure}[t]
    \centering
    \includegraphics[width=\linewidth]{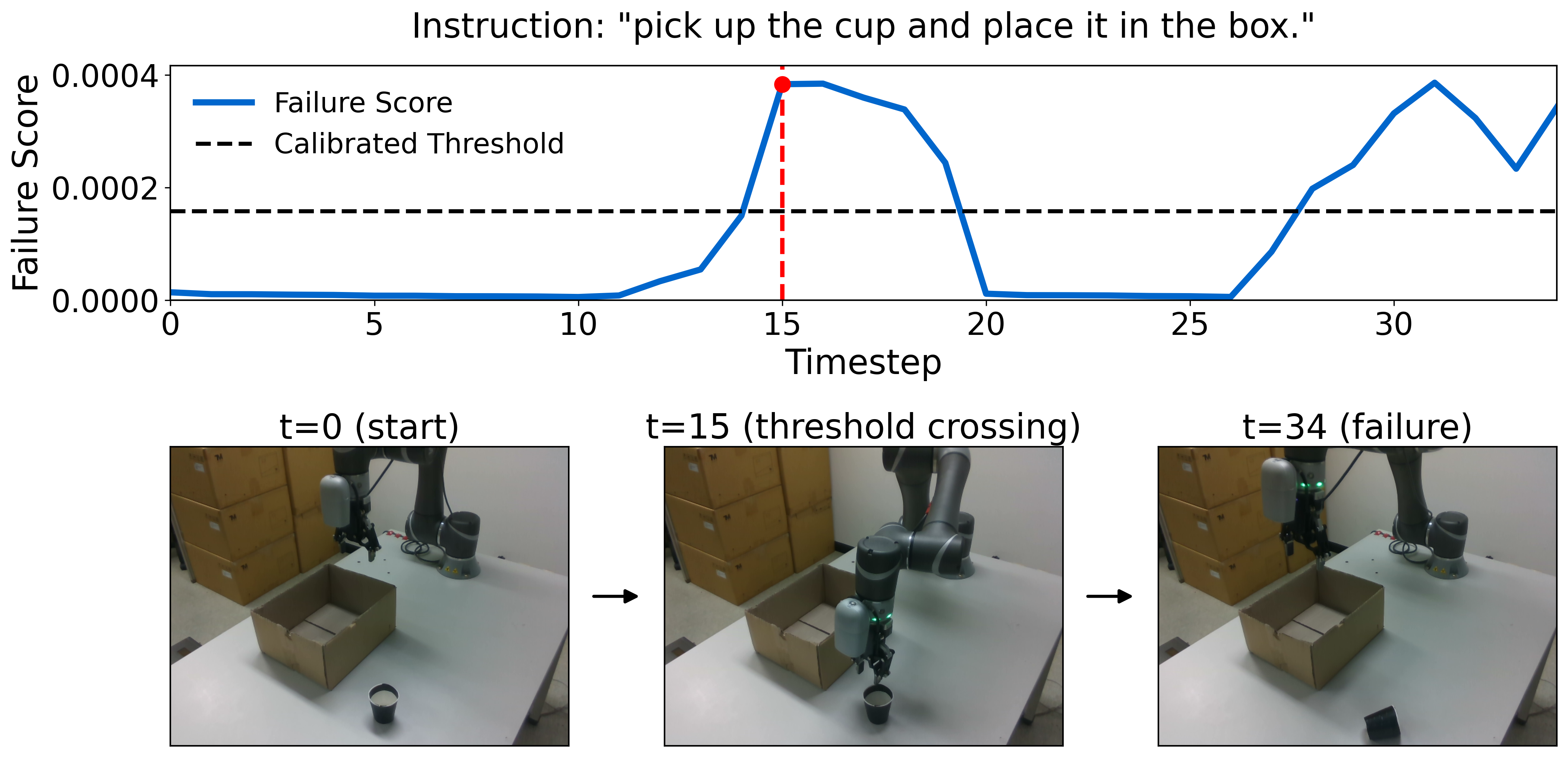}
    \caption{
    Representative real-world failure under the unseen black-cup object shift.
    The \method{} failure score crosses the calibrated threshold before the task failure. Bottom: observations at the start, first threshold
    crossing, and final failure. 
    }
    \label{fig:tm5_qualitative}
    \vspace{-2mm}
\end{figure}

\begin{table}[t]
    \centering
    \footnotesize
    \caption{
    Real-world sensitivity to perturbation strength on the ablation set,
    which comprises 20 successful and 10 failed rollouts under the ID
    green-cup condition ($K=2$).
    }
    \label{tab:real_sigma_ablation}

    \setlength{\tabcolsep}{10pt}
    \renewcommand{\arraystretch}{0.95}
    \begin{tabular}{cc}
        \toprule
        \textbf{Perturbation strength $\sigma$} & \textbf{AUROC} \\
        \midrule
        0.05 & 0.720 \\
        \textbf{0.10} & \textbf{0.750} \\
        0.15 & 0.740 \\
        0.20 & 0.655 \\
        0.30 & 0.400 \\
        \bottomrule
    \end{tabular}

    \vspace{1mm}

    \caption{
    Real-world failure detection under an object shift
    with 30 successful and 20 failed rollouts ($K=5$).
    Best and second-best are shown in bold and underlined.
    }
    \label{tab:real_robot_results}

    \renewcommand{\arraystretch}{0.95}
    \setlength{\tabcolsep}{4pt}
    \begin{tabular}{lccc}
        \toprule
        \textbf{Method}
        & \textbf{AUROC} $\uparrow$
        & \textbf{Bal. Acc.} $\uparrow$
        & \textbf{Det. Time} $\downarrow$ \\
        \midrule
        PCA-KMeans
        & 0.822
        & 0.733
        & 50.67 \\
        RND
        & \textbf{0.948}
        & \textbf{0.908}
        & 60.28 \\
        ACE
        & 0.716
        & 0.725
        & \textbf{33.41} \\
        Sampling
        & 0.803
        & 0.800
        & \underline{47.98} \\
        \method{}
        & \underline{0.862}
        & \underline{0.825}
        & 51.48 \\
        \bottomrule
    \end{tabular}
    \vspace{-2mm}
\end{table}

\subsection{Real-World Validation}
\textbf{Experimental Setup.}
We validate \method{} using a TM5-900 robot arm equipped with an
AG95 gripper and a fixed Intel RealSense D435i camera.
The camera captures $640\times480$ RGB images, which are provided to the fine-tuned $\pi_{0.5}$ policy together with the robot state, consisting of six absolute joint angles and one gripper value.
The language instruction is to ``pick up the cup and place it in the box.''

\textbf{Policy and Data Collection.}
We collect 50 demonstration episodes under the ID green-cup condition
and fine-tune $\pi_{0.5}$ for 10,000 optimization steps with an
effective batch size of 16.
The policy predicts 7-dimensional actions with an action chunk length
of 16, and all actions in each chunk are executed before replanning.
Camera observations are synchronized with the corresponding robot states
at 30\,Hz.

\textbf{Evaluation Protocol.}
Following the protocol in Sec.~\ref{sec:evaluation_protocol}, we construct four disjoint real-world rollout sets.
The reference and calibration sets each contain 20 successful
green-cup rollouts, with distinct rollouts used for baseline fitting and conformal threshold calibration.
The \textbf{ablation set} contains 30 green-cup rollouts, comprising 20 successes and 10 failures, and is used exclusively to select the perturbation strength $\sigma$.
The \textbf{test set} contains 50 rollouts, comprising 30 successes and 20 failures, all collected using an unseen black cup to evaluate an object shift from the ID green-cup condition.
A rollout is labeled as a failure if the task is not completed within
35 replanning steps or is terminated early for safety.
All methods are evaluated offline on the same cached policy inputs.

\textbf{Perturbation-Strength Ablation.} We use $K=2$ for the perturbation-strength sweep and $K=5$ for the final evaluation. AUROC remains stable for $\sigma\in[0.05,0.15]$ but degrades at $\sigma\geq0.2$. We therefore select the best-performing value, $\sigma=0.1$, for the real-world evaluation.

\textbf{Failure-Detection Results.}
Table~\ref{tab:real_robot_results} reports failure-detection
performance under the unseen black-cup object shift.
RND achieves the best performance in this setting, possibly because
representation-level novelty is more strongly correlated with task failure
in the relatively constrained single-task setup.
\method{} achieves an AUROC of 0.862 and a balanced accuracy of 0.825,
ranking second in both metrics and outperforming the action-level uncertainty
baselines, ACE (0.716/0.725) and Sampling (0.803/0.800).
Fig.~\ref{fig:tm5_qualitative} further shows a representative
real-world failure in which the \method{} score crosses the calibrated
threshold before the eventual task failure.
Overall, these results show that \method{} remains competitive for
real-world VLA failure detection.

\section{Conclusion}
In this work, we proposed \method{}, a training-free failure detection
framework for pretrained VLA models that estimates epistemic uncertainty
from disagreement induced by low-rank weight perturbations.
Experiments on LIBERO-PRO show that \method{} achieves the highest average AUROC and balanced accuracy among the evaluated methods while
consistently outperforming stochastic action sampling across distribution shifts.
Real-world evaluation further demonstrates competitive performance,
with \method{} ranking second in both AUROC and balanced accuracy.
Although \method{} remains effective with a small number of
perturbation samples, multiple perturbed forward passes still introduce additional inference cost.
Future work could reduce this overhead through more efficient
uncertainty estimation or distillation-based approximations.
Overall, our results suggest that perturbation-based epistemic uncertainty is a promising direction for reliable failure detection in VLA systems.

\bibliographystyle{IEEEtran}
\bibliography{references}

\end{document}